\documentclass[times,twocolumn,final,authoryear]{elsarticle}
\usepackage{prletters}
\usepackage{framed,multirow}
\usepackage{amssymb}
\usepackage{latexsym}
\usepackage{graphicx}
\usepackage{amsmath}
\usepackage{adjustbox}
\usepackage{subfig}
\usepackage{cases}
\usepackage{algorithm}
\usepackage{algpseudocode}
\usepackage{pifont}
\usepackage{multirow}
\newcommand{\cmark}{\ding{51}}%
\newcommand{\xmark}{\ding{55}}%
\usepackage{url}
\usepackage{xcolor}

\begin{document}
\begin{frontmatter}

\tnotetext[t1]{\textit{This paper is under review to "Pattern Recognition Letters"}}
\tnotetext[t2]{Code \textit{https://gitlab.com/artelabsuper/ocdmst}}

\title{Dynamic Decision Boundary for One-class Classifiers \\ applied to non-uniformly Sampled Data}

\author[1]{Riccardo La Grassa} 
\author[1]{Ignazio Gallo}
\author[1]{Nicola Landro}
\address[1]{University of Insubria, Department of Theoretical and Applied Science, Varese, Italy}

\begin{abstract}
\textit{A typical issue in Pattern Recognition is the non-uniformly sampled data, which modifies the general performance and capability of machine learning algorithms to make accurate predictions. 
Generally, the data is considered non-uniformly sampled when in a specific area of data space, they are not enough, leading us to misclassification problems. This issue cut down the goal of the one-class classifiers decreasing their performance. 
In this paper, we propose a one-class classifier based on the minimum spanning tree with a dynamic decision boundary (OCdmst) to make good prediction also in the case we have non-uniformly sampled data.
To prove the effectiveness and robustness of our approach we compare with the most recent one-class classifier reaching the state-of-the-art in most of them}.
\end{abstract}
\begin{keyword}
One-class classifier\sep Machine Learning\sep Covid-19 dataset
\end{keyword}
\end{frontmatter}

\section{Introduction}
In the deep learning era, the best models need to have many hundreds or thousands of instances to make good predictions and to reduce as little as possible misclassification. 
However, in the scenario where we have few samples of a specific class without knowing anything about other classes, the neural network models can not be used.
In this case, different approaches exist that uses neural networks for this task (\cite{perera2019learning}). 
They train the model using a dataset of a different context than the one in which the model will then be used.
Although it offers good results in terms of classification, it does not be applied to a specific context, for instance, in case the source of the test set is very different from the data of the pre-trained model and consequently, the discrimination power model will be very low to generalize these test instances.
In conjunction with few objects available and time required to train a model with a huge dataset, deep models might not be the right choice for this specific task.
In \cite{grassa2020ocmst}, we used a neural network pre-trained on a large dataset as Imagenet to extract deep features and use them as input for a one-class classifier (OCC) based on MST (OCmst). 
They introduce another level of boundary used to label all uncertain images and construct another one-class MST using all negative samples classified. The final step will classify all uncertain value starting from a couple of MST. 
Although the idea is interesting they do not handle dynamic boundaries but only different levels of static boundaries are treated.
In literature, various approaches exist for one-class classification. 
For example, \cite{tax2002using} and \cite{Krawczyk} use ensemble methods to find best partitions or to handle under/over-sampling in an unbalanced scenario. 
A classical SVM revised algorithm (OCSVM) (\cite{scholkopf2001estimating}) offers good results with different kinds of data. 
Other interesting models are graph-based (\cite{pekalska},~\cite{la2019classification},~\cite{la2019binary},~\cite{liu2016fast_ref}) and occur on a different domain as computer vision (\cite{perera2019learning},~\cite{ ruff2018deep}), bioinformatics (\cite{pekalska},~\cite{zhang2014one}), clustering (\cite{liu2016fast_ref}).
Furthermore, in the literature other well-known models exist, like hybrid models.
In this scenario, neural network models are used as deep features extractors jointly to classical machine learning classification algorithms like a one-class support vector machine.~\cite{scholkopf2001estimating}, autoencoder+ocsvm \cite{andrews2016detecting}, autoencoder plus knn \cite{song2017hybrid}, autoencoder plus svdd \cite{kim2015deep}.
Intuitively, the main advantages of using hybrid models are the reduced number of instances and to exploit the power of neural network models to provide the best discriminative features. 
In \cite{chalapathy2018anomaly} and \cite{ruff2018deep}, the authors propose an approach to combine Convolutional Neural Networks (CNN) with SVDD, while optimizing the class boundary in the output features space.
Some disadvantages of this methodology are the computational time required for the training step and in the scenario which we have few data this approach might not have good results.
In this paper, we propose an extension of our previous works \cite{la2019classification} and \cite{la2019binary}, using a new approach based on inverse sigmoid function to handle the decision boundary in a dynamic methodology. 
We conduct a wide variety of experiments on different datasets to demonstrate the rightness of our model and compare it with the most recent one-class classifiers, overcoming them in many tasks.
Today, finding good models in particular situations such as in medicine is a necessity.
We tested our model on Covid-19 disease using clinical images from \textit{\cite{radiopedia}} and \textit{\cite{sirm},} obtaining interesting results.
In the following sections we describe the main approach using a minimum spanning tree as class descriptor \ref{sec:one-class-classifiers}, we introduce our approach \ref{sec:our_approach} and finally we will discuss our results compared with other known models \ref{sec:experiments}.

\section{One-class Classifiers}\label{sec:one-class-classifiers}
One-class novelty detection refers to the recognition of abnormal patterns on data recognized as normal.
Abnormal data, also know as outliers are patters who belong to the different classes than a normal class.
The goal of the novelty detection field is to distinguish anomaly patterns that are different from normal and classify them.
The capability of many machine learning technique, in the field of one-class classification is to decide whether a new instance belongs to the same class of data or if it has different behavior such as to be considered as an outlier.
Designing an OCC classifier is not a simple task due to the nature of data available and from the quality of data sampled. In Tab.~\ref{tab:capabilities_oc} we show the general capabilities of one-class classifiers.

\begin{table}[t]
\caption{\textbf{Capabilities of state-of-the-art OCC algorithms on target class features. Our model uses a dynamic boundary for the target class shape, differing from other models that make use of a static boundary around the class.}}
\label{tab:capabilities_oc}       
\begin{adjustbox}{width=\columnwidth,center}
\begin{tabular}{l|cccc}
\hline\noalign{\smallskip}
Models & multi-modality & multi-density & noise & Shaped on target class \\
\noalign{\smallskip}\hline\noalign{\smallskip}
Gaussian estimator  &\xmark &\xmark &\xmark &\xmark  \\
MoG       &\cmark  &\cmark &\xmark &\xmark           \\
MST\_CD  &\xmark &\xmark &\xmark &\cmark \\
OCmst  &\xmark &\cmark &\xmark &\cmark \\
OC-SVM (RBF) &\cmark  &\xmark &\cmark  &\cmark \\ 
SVDD (RBF) &\cmark  &\xmark &\cmark  &\cmark \\
OCdmst (our) &\xmark  &\xmark &\xmark  &Dynamic \\
\noalign{\smallskip}\hline
\end{tabular}
\end{adjustbox}
\end{table}

In Tab.~\ref{tab:capabilities_oc} the data has multi-modality and exists more than one distribution in it \cite{hovelynck2010multi}.
The general approach of multi-density is not a simple task. 
An object might have many levels of significance in the proximity of decision boundary showing that a single density level in some cases can be reductive in terms of classification.
The noise in the data can lower the general performance of classifiers lead it to misclassification.
Arbitrarily shaped distribution of the target class is a common feature of the MST, SVDD or MoG models, that are able to cover the data according to their methodology.

In \cite{liu2016modular}, the authors propose an interesting approach based on density modular ensemble classifiers for the one-class classification task.
They applied a pruning method to split the original structure of an MST, built from the target class while creating different clusters.
In this work, no cluster algorithms are applied.
They used a Gaussian estimator algorithm to build a decision border for each cluster created in the previous step.
Parameters considered are the level of the pruning step and all parameters behind the one-class chosen for each partition built.
The authors describe the small density or isolate nodes as noises and remove them in the pruning step including only all samples of the dense region extracted by local dense subset method. 
The local dense subset can capture dense small regions of the target sample excluding all data far away from it.
In \cite{liu2016fast_ref}, authors use the k-means clustering algorithm   in the given target class. Then, a minimum spanning tree is build considering all the clusters centroids computed before. Furthermore, they use the second partitioning step to smooth the boundary because gaps through the clusters exist. 
Finally, a Gaussian estimator or an MST Class Descriptor (MST\_CD) is applied to build a decision boundary for each cluster.
Both works show us optimal results in terms of accuracy and overcome the state-of-the-arts in many datasets from the UCI repository.
However, the following issues exist:
-- a bad choice of several clusters can be a problem that can lead to misclassification; 
-- isolates nodes or medium-large density of nodes are avoided because considered as outliers.
This solution could cause problems in the decision boundary since in the case of non-uniformly sampled data the density of the region can be varied by noisy samples removal.

In \cite{tax2004support}, using SVDD (Support Vector Data Description), the authors describe some models related to OCSVM, where a hypersphere is considered to separate the data instead to use a hyperplane. The goal is to find the smallest hypersphere with center $c \in F_k$ and radius $R >0$ that encloses as much as possible data in feature space $F_k$
Arbitrarily shaped distributions are showed in our previously works (\cite{la2019classification},~\cite{la2019binary}). 
Here the main idea is to cut down the computation of the minimum spanning tree and to avoid data noise using only a small partition of the target class as a class descriptor. Therefore, the method is iterated for each test to classify. 
The decision boundary built is different for each run, because we construct an MST starting from the neighbors of each test sample to be classified.
However, the approach offers good results, the decision border remains the same in all the neighbors, also in regions with a small density. This problem can lead us to expand the boundary too much in all regions due to the low density of non-uniformly sampled data.
Starting from \cite{la2019binary}, we elaborated our model with a dynamic approach to choose the boundary based on the density of the nearest region to the test instance. In the following section, we discuss the main approach using an MST as a class descriptor and then introduce our proposed model.

\section{The main approach using a Minimum Spanning Tree}
Many task predictions have obtained successful results using specific structure in Graph Theory as spanning trees and have found many application contexts in social network (\cite{wang2019hierarchical}), biology (\cite{isaza2018biological}), companies and political events (\cite{memon2020general}). 
In this section, we focus on the minimum spanning tree used as a class descriptor and highlight the pro and cons of this model. This brief overview is useful to understand our approach and a possible solution to increase the performance of this classifier.
In general, MST\_CD model uses the following two alternatives to measure the distance of a point $x$ from the graph:
\begin{itemize}
    \item projection $p_{e_{i,j}}(x)$ of a point $x$ on an edge $e_{i,j}$= ($x_i$,$x_j$), defined as follow
    \begin{equation}
    p_{e_{i,j}}(x)=x_i + \frac{(x_j - x_i)^T(x - x_i)}{||x_j - x_i||^2}(x_j - x_i)
    \end{equation}
    \item $\min(||x - x_j||,||x - x_i||)$.
\end{itemize}

If $p_{e_{i,j}}(x)$ lies on the edge $e_{i,j}$= ($x_i$,$x_j$), we compute $p_{e_{i,j}}(x)$ and the Euclidean distance between $x$ and $p_{e_{i,j}}(x)$, more formally:
\begin{equation}
0<=\frac{(x_j - x_i)^T(x - x_i)}{||x_j - x_i||^2}<=1
\end{equation}
then
\begin{equation}
(x|e_{i,j})=||x - p_{e_{i,j}}(x)||
\end{equation}
Otherwise, we compute the Euclidean distance of $x$ and the two points $x_i$ and  $x_j$. More precisely:
\begin{equation}
d(x|e_{i,j})=min(||x - x_j||,||x - x_i||)
\end{equation}

\begin{table}
\centering
\caption{\textbf{Performance comparison of OCdmst and Resnet18 on Covid-19 dataset. We highlight the capability of OCdmst to make prediction even when the neural network does not recognize values from negative class. Target class used is Positive label.}}
\label{tab:covid_results}       
\begin{tabular}{l|ccc}
\noalign{\smallskip}\hline\noalign{\smallskip}
K-Fold & Positive class & & Negative class\\
\hline
& & \textbf{OCdmst} \\
\noalign{\smallskip}\hline\noalign{\smallskip}
1  & \textbf{0.853} & & \textbf{0.347} \\
2  & \textbf{0.843} & & \textbf{0.187}  \\
\hline
 & & \textbf{Resnet18} \\
\hline
1 & 0.781 & & 0 \\
2 & 0.828 & & 0 \\
\noalign{\smallskip}\hline
\end{tabular}
\end{table}

Then, a new instance $x$ is recognized from a MST if it lies within the boundary, otherwise, the object is considered an outlier.
In \cite{pekalska}, \cite{la2019binary}, \cite{la2019classification} the decision whether an object is recognized by classifier or not is based on a threshold $\theta$ applied to the shape created in this phase. More formally:
\begin{equation}
d(x|X) \leq \theta
\end{equation}
The threshold $\theta$ is a parameter used to assign the boundary dimension of a spanning tree. 
Then, given $\hat{e}=(||e_1||,||e_2||,...,||e_n||)$ as an ordered set of edges weights, we define $\theta$ as 
\begin{equation}\label{eq:static_thr}
    \theta = ||e_{[\alpha n]}||,\text{where } \alpha \in [0,1]
\end{equation}{}
For instance, with $\alpha=0.5$, we assign to $\theta$ the median value of all edges weights of the spanning tree.
The curse of the number of instances is a possible weakness that is reflected in an increase in terms of computational time, it also does not handle non-uniformly sampled data. This lack results in a too-large decision border in the whole area with few data.

\begin{figure}
    \centering
    \includegraphics[width=30mm,height=30mm]{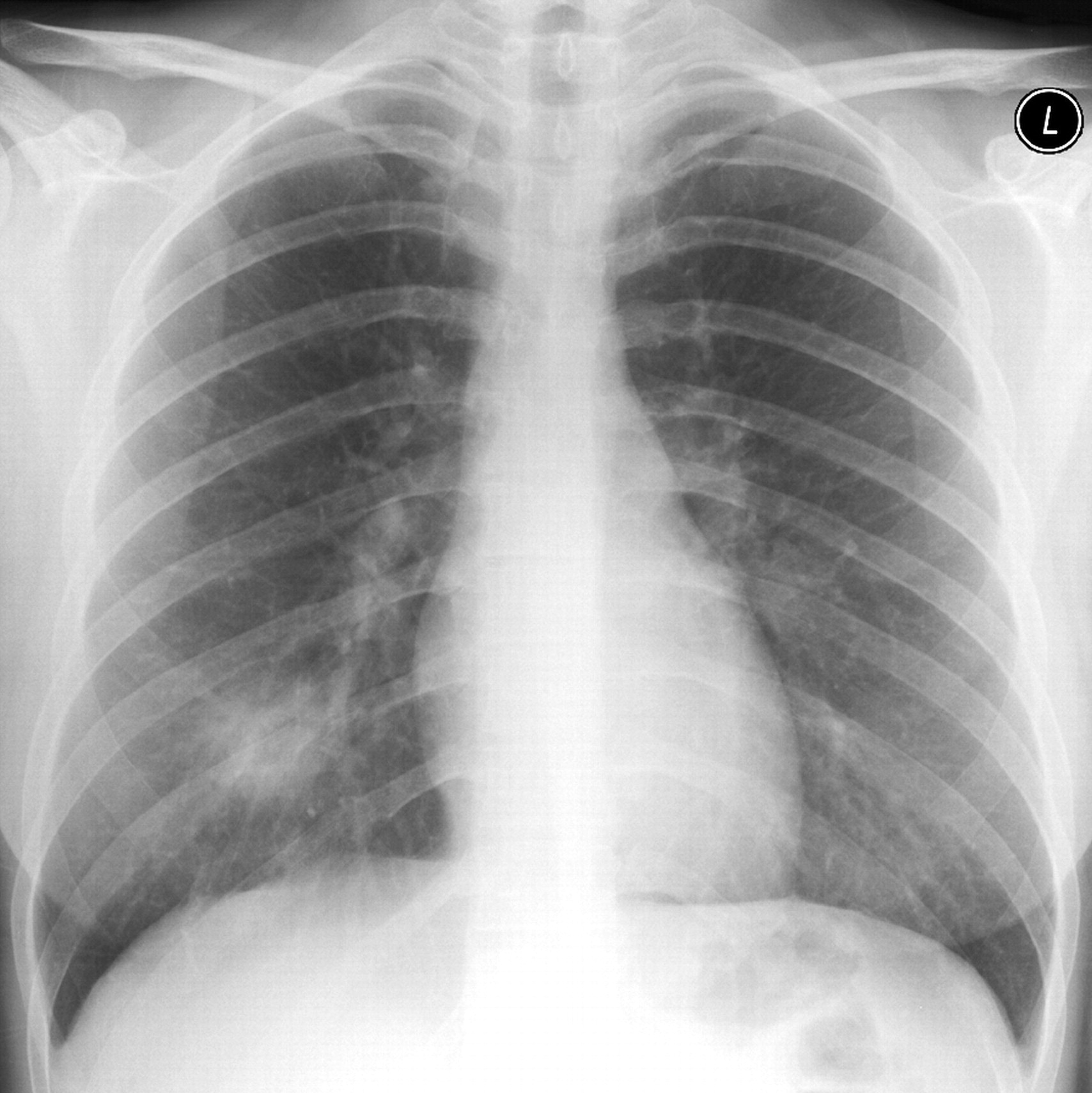}
    \includegraphics[width=30mm,height=30mm]{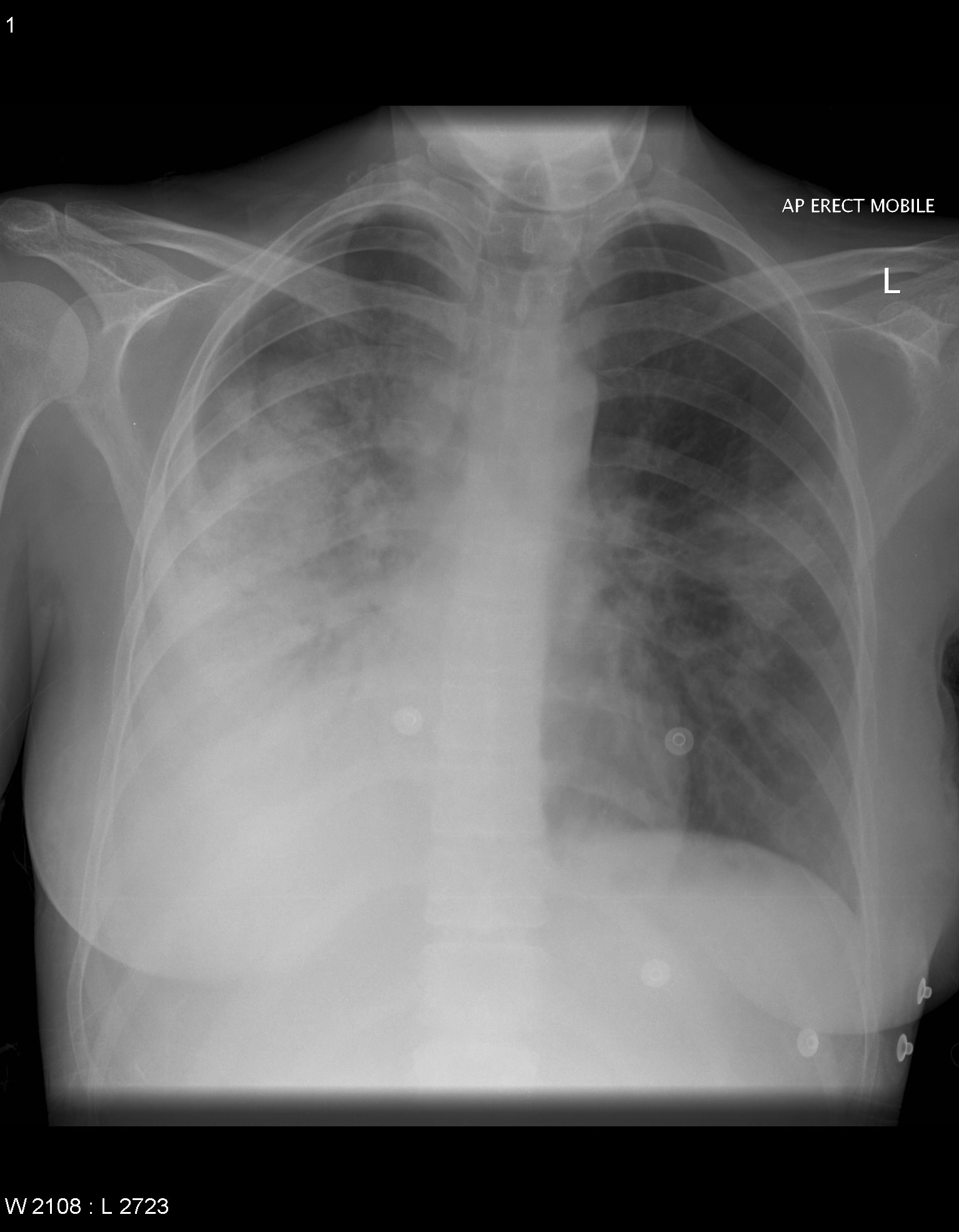}\\
    \includegraphics[width=30mm,height=30mm]{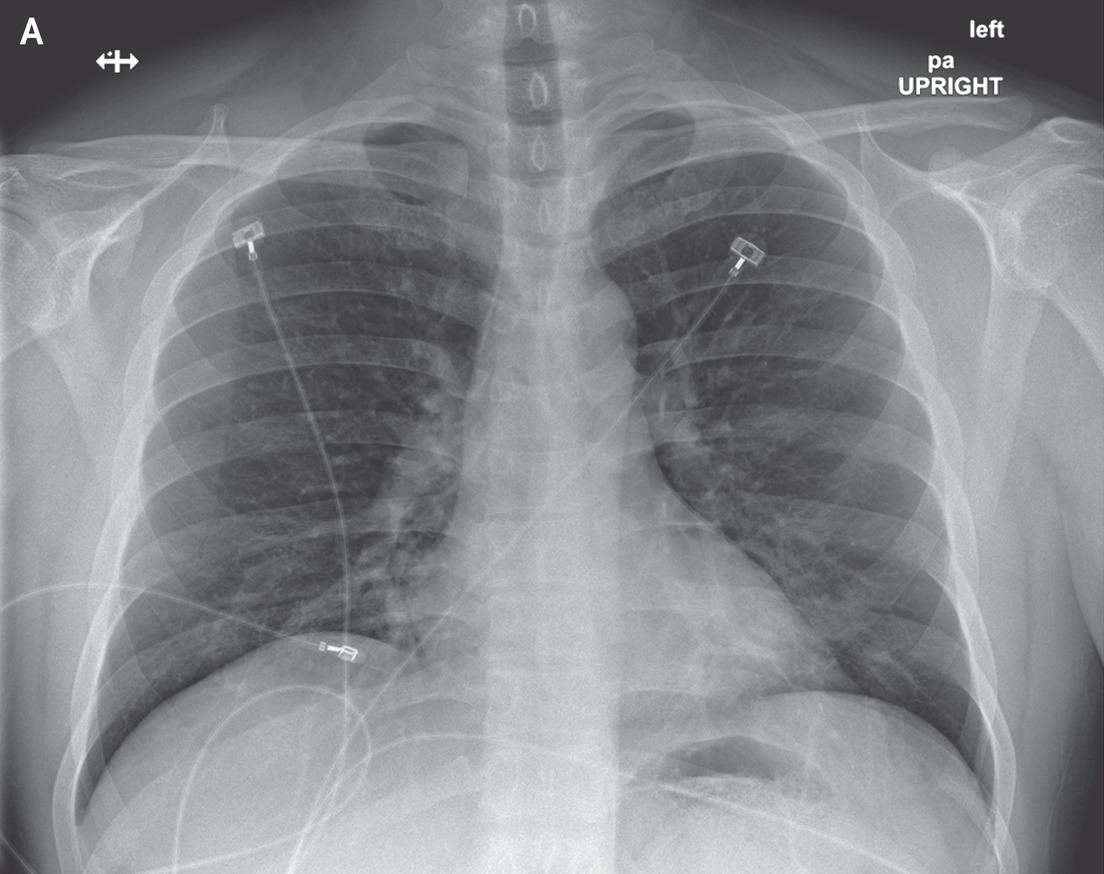}
    \includegraphics[width=30mm,height=30mm]{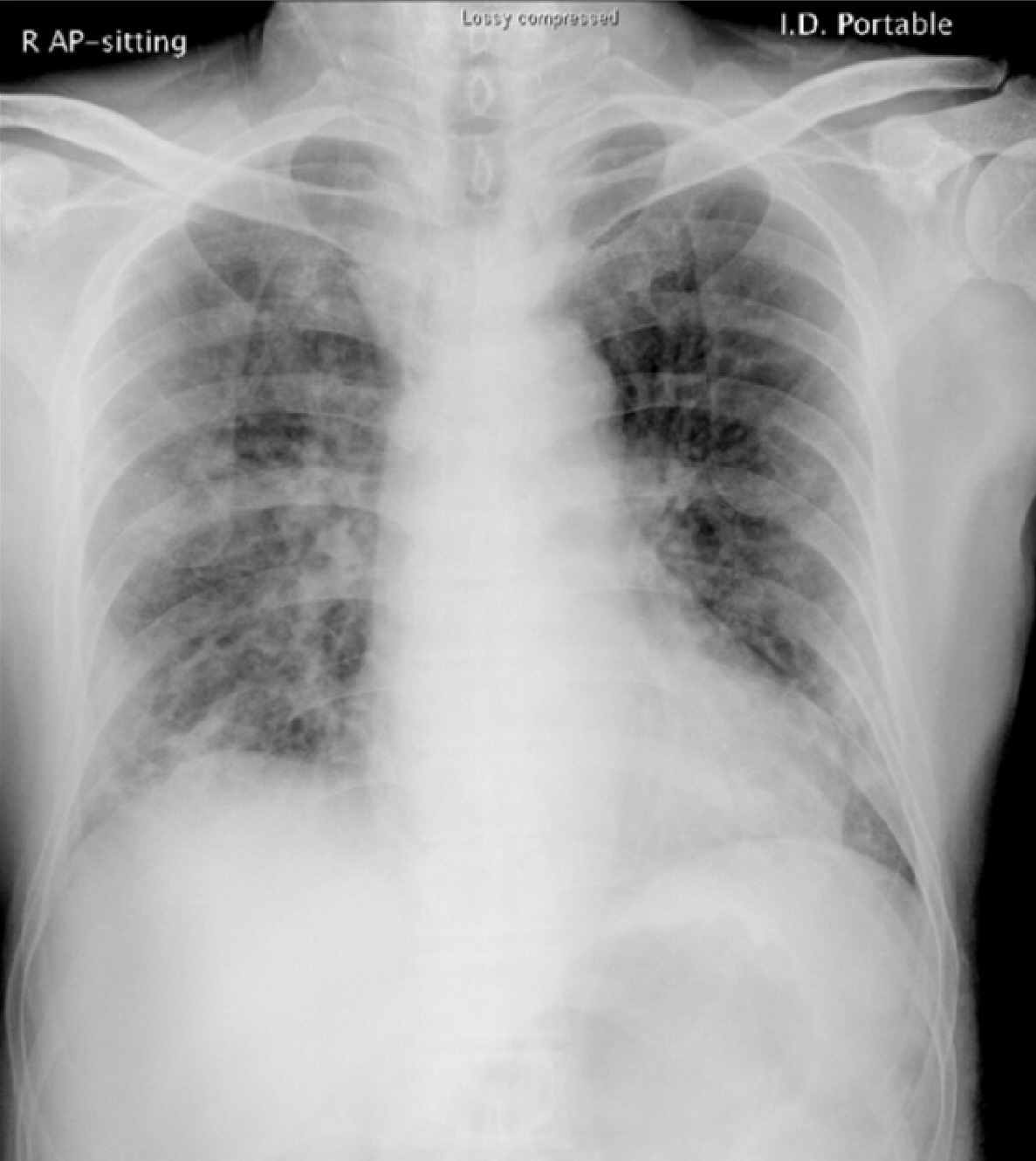}
    \caption{\textbf{Four images extracted from Covid-19 dataset.
    Figs (a) and (b) represents negative cases with pneumonia and SARS respectively, Figs (c) and (d) are positive cases of Covid-19.
    Due to different variety of images we convert all images to gray scale and re-scaled to $256\times 256$ pixels.}} 
    \label{fig:covid-19_dataset}
\end{figure}

\begin{figure}[t]
    \centering
    \includegraphics[width=.45\textwidth]{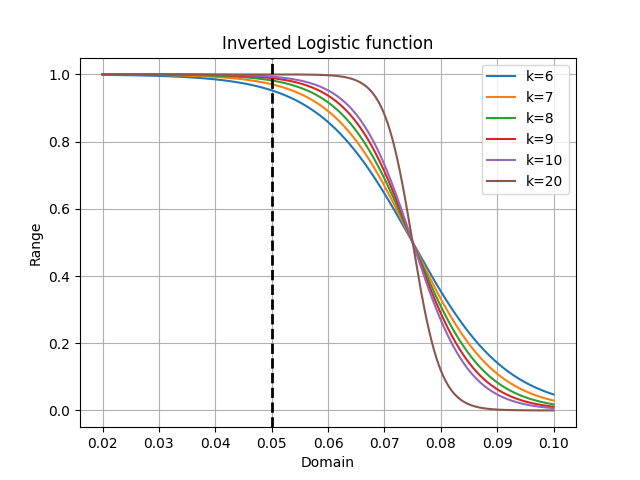}
    \caption{\textbf{Logistic function with grow rate $k \in \{6,7,8,9,10,20\}$. 
    For all experiments we chose $\beta=1.5$ to move the inflection point of the curve with respect to the median value of all the standard deviations starting from $N$ random samples in small MST built.}}
    \label{fig:logistic_function}
\end{figure}

\section{Our approach}\label{sec:our_approach}
\begin{figure}
    \centering
    \includegraphics[width=1\columnwidth]{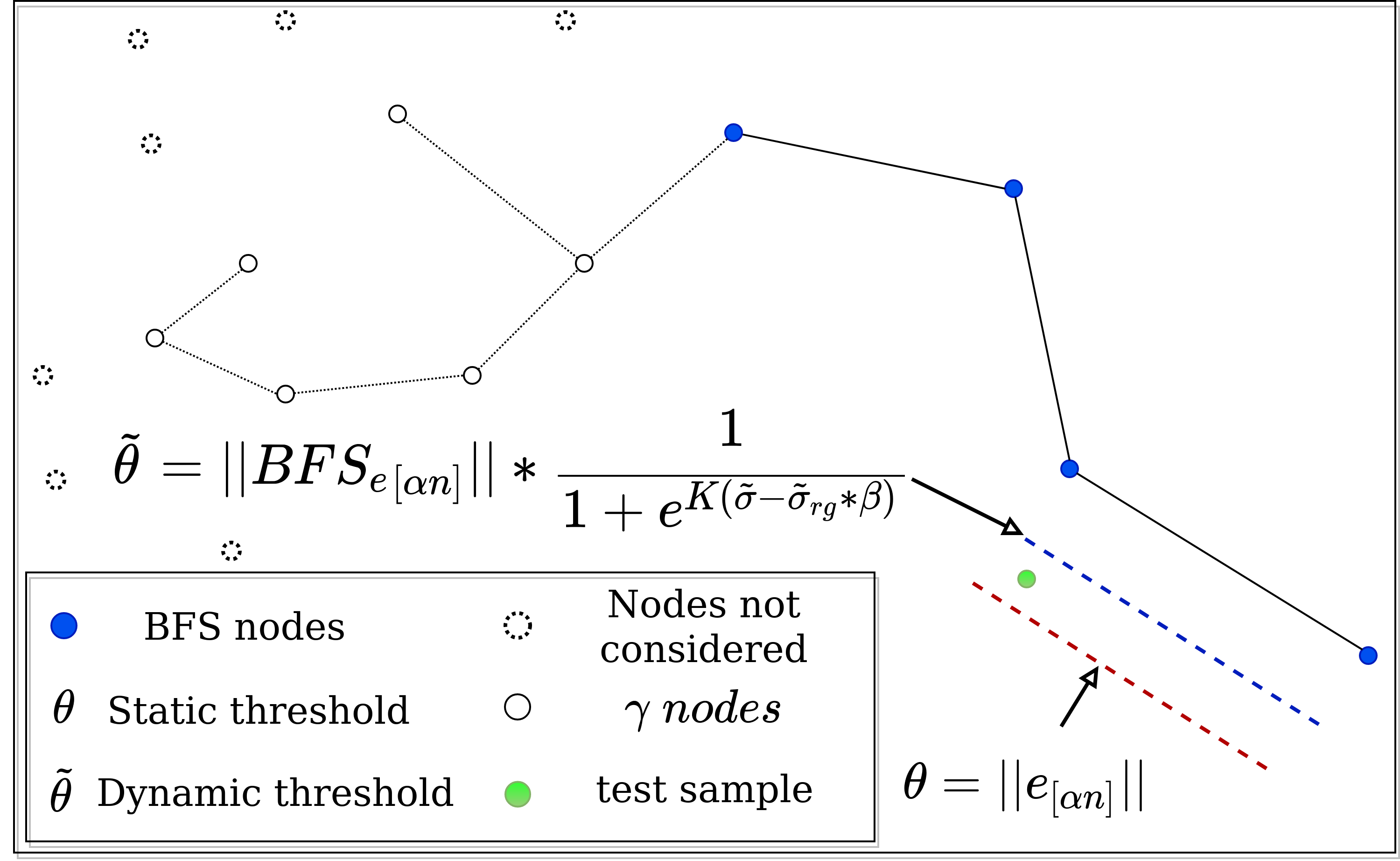}
    \caption{\textbf{Overview of OCdmst algorithm. 
    This figure emphasizes the dynamic border relationship with the nearest instances.}}
    \label{fig:ocdmst_case}
\end{figure}

Starting with our previous works \cite{la2019binary} and \cite{la2019classification} we use the same baseline called MST\_CD\_gp. This last approach is different from \cite{pekalska} because it uses a couple of small MST built for each test instance to make a prediction and it handles the case in which both classifiers has right or wrong. To remain in the context of a one-class classifier, we use only a target class of size $\gamma$ to make predictions. 
In previous papers, we catch the structure of the target class using MST and built a decision boundary $\theta$ using Eq. \ref{eq:static_thr}. 
However, authors do not consider the problem of non-uniformly sampled data. 
In specifics regions of the euclidean space, it is possible to found the sparse zone and consequently to obtain a good boundary from the target class (see red line in Fig.\ref{fig:ocdmst_case}) can become a problem. To remedy this problem we need to build a general methodology to create a dynamic decision boundary and avoid the wrong threshold due to sparse data.
For each test instance, we search the first nearest $\gamma$ nodes and use the Kruskal algorithm to build an MST\_CD\_gp, then we simply apply a Breadth-first search starting to the nearest object to test instance. In this way we preserve the structure around a target class created by an MST of size $\gamma$ and takes $N$ nodes by BFS search, more formally, given $\hat{e}=(||e_1||,||e_2||,...,||e_n||)$ be a set of weighted edges extracted by a BFS search algorithm applied on the MST\_CD\_gp, we introduce a dynamics threshold $\tilde{\theta}$ as:
\begin{equation}\label{eq:dynamic_threshold}
    \tilde{\theta} = ||BFS_e {_{[\alpha n]}}|| * \frac{1}{1+e^{K(\tilde{\sigma}-\tilde{\sigma}_{rg}*\beta)}}
\end{equation}{}
where $\alpha=0.5$ in order to take the median of all the edges weights, $\hat{\sigma}$ is the normalized standard deviation of all nodes selected by BFS with variance computed with his data instances distribution (see blue line in Fig.\ref{fig:ocdmst_case}) and $K$ is the logistic growth rate or steepness of the curve (we used $K=5$).
$\tilde{\sigma}_{rg}$ represents the median value of all standard deviation computed taking $N$ random group having the size returned by the $BFS$ with depth $d$. This last is an important value that defines the inflection point of our inverse sigmoid. 
Furthermore, we use a variable $\beta$ useful to shift the inflection point and thus we slow the fall to 0 (see Fig.~\ref{fig:logistic_function}).

Note that our mean value of standard deviation will be the centroid computed considering all target class instances.
From Popoviciu's inequality (\cite{popoviciu1935equations}), it is well-known that the upper bound of the variance of any random variable $X$ on a range $[x_{min}, x_{max}]$ is equal to:
\begin{equation}\label{upperbound}
    V[X] \leq {\frac{(x_{max}-x_{min})^2}{4}}
\end{equation}{}

Since we handle instances $x_i$ with dimension $d > 1$, the deviation standard is:
\begin{equation}
\sigma = \sqrt {\frac{\sum_{i=1}^{n}(x_i - \bar{x})^2} {d*n} }
\end{equation}
Keep in mind the relationship in Eq. \ref{upperbound}, the upper bound of $\sigma$ will be:
\begin{equation}
    \sigma \leq {\sqrt{\frac{(x_{max}-x_{min})^2}{4}}}
\end{equation}{}

Therefore in Eq. \ref{eq:dynamic_threshold} we normalize $\sigma$ as 
\begin{equation}\label{std_normalize}
    \hat{\sigma}=\frac{\sigma}{\sqrt{\frac{(x_{max}-x_{min})^2}{4}}}
\end{equation}{}
where $x_{max}$ and $x_{min}$ are the maximum/minimum range of our data.

\begin{figure}
    \centering
    \includegraphics[width=80mm,height=60mm]{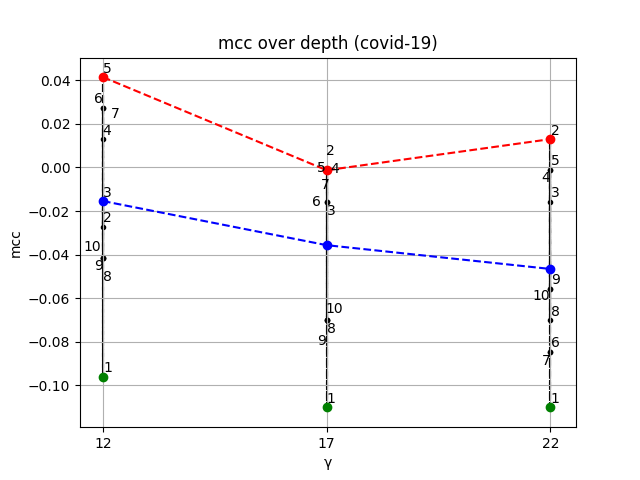}
    \caption{\textbf{Analysis of OCdmst using different size of $\gamma$ on different levels of depth $d$ on Covid-19 dataset. 
    In most of the datasets used we obtain better results using BFS with low $depth$. 
    In this case the target class used is the positive label.}} 
    \label{fig:gamma_analysis_covid}
\end{figure}

Our approach leverages the distribution around the neighbors of test samples and uses metrics that are well-known in the literature. 
So, the main effect will be to deflate the decision boundary with high power in a non-uniformly sampled area of data and with low power in a uniformly sampled zone. 
We remember that we normalize $\sigma$ in Eq. \ref{std_normalize} using the upper bound from Popoviciu's inequality.
This simple approach allows us to shrink the decision boundary in the sparse data zone and to decrease with less power his area near the data with a low standard deviation.
Our approach can be summarized as below:
\begin{enumerate}
    \item Build a small MST using the first $\gamma$ instances nearest to the test sample.
    \item Apply BFS with parameter depth $d$ on the small MST constructed before and return $\sigma$ and the median of the weighted edges obtained from BFS search.
    \item Take the median of N random nodes of small MST of size $d$. This value will be the inflection point of our inverse sigmoid function (all standard deviation are normalized \ref{std_normalize}).
    \item Using threshold extracted by step 2 shifted by $\beta$ value and build a dynamic decision boundary for each instance, then apply the classical MST\_CD\_gp model described in \cite{la2019binary}.
\end{enumerate}{}

\begin{algorithm} \caption{\textbf{OCdmst}}
\label{alg:ocdmst}
\scriptsize
\begin{algorithmic}[1]
\State{$G_0$=\text{Target class}}
\State{$x$=\text{All test instances}}
\For {$v \in \mathcal G_0 $}
\State all euclidean distances $\gets || x - v ||$
\EndFor
\State Create a small mst from $\gamma$ neighbors from test x
\State NodeX = Take min(all euclidean distances) and return node $v$
\State EdgesNodeX $\gets$ Search inc/out edge nodeX and return $(x_i, x_j)$
\For {$x_i,x_j \in EdgesNodeX$ }
\If {$0<=\frac{(x_j - x_i)^T * (x-x_i)}{||x_j - x_i||^2}<=1$}
\State {$P_{e_{_{i_{j}}}}(x)=x_{i} + \frac{(x_{j} - x_{i})^T * (x-x_i)}{||x_j - x_i||^2}*(x_j-x_i)$}
\State {$d(x|e_{_{i_{j}}}) \gets ||x - P_{e_{_{i_{j}}}}(x)||$}
\Else
\State{$d(x|e_{_{i_{j}}}) \gets min \big\{ ||x - x_i||, ||x -x_j||\big\}$}
\EndIf
\EndFor
\State{$\sigma, \tilde{\mu} \gets \text{BFS search from small mst with depth}=d$}
\State{$\tilde{\sigma}_{rg} \gets \text{median value from N random nodes of small mst with size of BFS}$}

\State{All $\sigma$ normalization (see \ref{std_normalize})}
\State{$\tilde{\theta} = ||BFS_e {_{[\alpha n]}}|| * \frac{1}{1+e^{K(\tilde{\sigma}-\tilde{\sigma}_{rg}*\beta)}}$}
\State {1 $\gets$  \text{if  $d_{MST\_CD_0}(x|X) <= \tilde{\theta}$}} 
\State {0 $\gets$ \text{if  $d_{MST\_CD_0}(x|X) >  \tilde{\theta}$}}
\end{algorithmic}
\end{algorithm}

\begin{figure}
    \centering
    \includegraphics[width=.3\textwidth]{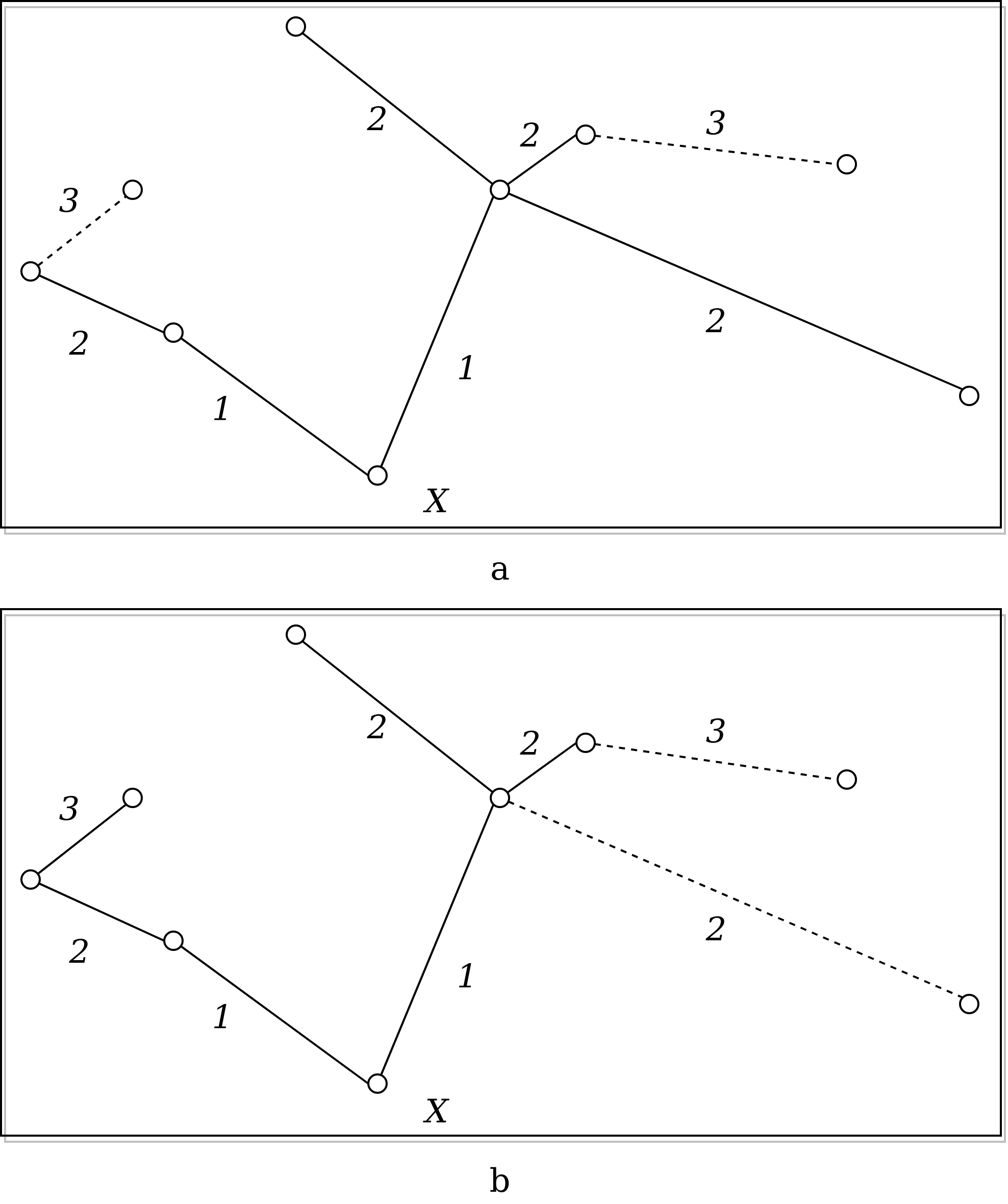}
    \caption{\textbf{In (a) the BFS path results starting from the node closest to the $X$ test using $depth=2$ and $\gamma = 9$.
    In (b), the structure created without the use of BFS. 
    The difference is that in (a) we capture the neighbor's relations in terms of connectivity instead in (b) we use the Euclidean distance.
    }}
    \label{fig:ocdmst_bfs}
\end{figure}

\section{Experiments}\label{sec:experiments}
To prove the effectiveness and robustness of our OCdmst, we use the same experimental setting used in \cite{wang2019dynamic} and compute Matthews's Coefficient Correlation (MCC) and in some experiments we reported total accuracy as performance metrics.
In the binary classification problem, a model can obtain high accuracy on outliers class but very low performance to recognize target class or vice versa. 
In such a situation, using a global metric as the classical accuracy is not the right way to compute the classification score. 
To avoid this issue we use MCC as a metric that can be used even if we have classes with different sizes. 
MCC returns a value between -1 and +1 where a coefficient of +1 represents a perfect prediction, 0 random predictions and -1 indicates total disagreement between prediction and observation.
\begin{equation}
    {\displaystyle {\text{MCC}}={\frac {{\mathit {TP}}\times {\mathit {TN}}-{\mathit {FP}}\times {\mathit {FN}}}{\sqrt {({\mathit {TP}}+{\mathit {FP}})({\mathit {TP}}+{\mathit {FN}})({\mathit {TN}}+{\mathit {FP}})({\mathit {TN}}+{\mathit {FN}})}}}}
\end{equation}

\begin{table*}
\centering
\caption{\textbf{Performance comparison of OCdmst and state-of-the-art OCC methodology on UCI datasets using MCC metric. In bold our average result.}}
\label{tab:mcc_comparison}       
\begin{adjustbox}{max width=\textwidth}
\begin{tabular}{l|ccccccccc}
\hline\noalign{\smallskip}
Dataset & FS-EOCC & FS-EOCC & TOCC & OCClustE & OCClustE & Gauss & MST\_CD & OC-SVM & \textbf{OCdmst}  \\
        & (Gauss) & (MST\_CD) &    & (Gauss)  & (MST\_CD) &      &         &        &                   \\  
        
\noalign{\smallskip}\hline\noalign{\smallskip}
Breast (Benign)  & 0.765 (0.006) & 0.864 (0.003) & 0.847 (0.004) & 0.899 (0.001) & 0.905 (0.003) & 0.895 (0.0) & 0.894 (0.002) & 0.473 (0.002) & 0.774  \\
Breast (Malignant)  & 0.471 (0.022) & 0.218 (0.009) & 0.215 (0.012) & 0.208 (0.027) & 0.313 (0.009) & 0.19 (0.007) & -0.015 (0.011) & 0.234 (0.005) & 0.204  \\
Diabetes (Absent)  & 0.096 (0.008) & 0.005 (0.002) & 0.05 (0.003) & -0.01 (0.005) & 0.051 (0.004) & 0.009 (0.005) & -0.015 (0.004) & 0.010 (0.007) & 0.066  \\
Diabetes (Present)  & 0.186 (0.004) & 0.233 (0.008) & 0.209 (0.003) & 0.228 (0.004) & 0.21 (0.012) & 0.161 (0.016) & 0.206 (0.003) & 0.203 (0.016) & 0.178  \\
Glass (Float)  & 0.505 (0.017) & 0.567 (0.016) & 0.471 (0.012) & 0.452 (0.006) & 0.507 (0.005) & 0.43 (0.011) & 0.422 (0.005) & 0.389 (0.005) & 0.535  \\
Glass (NoFloat) & 0.375 (0.022) & 0.32 (0.003) & 0.314 (0.015) & 0.339 (0.024) & 0.336 (0.002) & 0.251 (0.015) & 0.267 (0.017) & -0.149 (0.035) & 0.238  \\
Heart (Present)  & 0.062 (0.029) & 0.095 (0.026) & 0.004 (0.017) & 0.015 (0.010) & 0.03 (0.011) & 0.01 (0.021) & 0.071 (0.005) & 0.069 (0.005) & 0.037 \\
Heart (Absent)  & 0.082 (0.037) & 0.231 (0.006) & 0.115 (0.011) & 0.095 (0.024) & 0.249 (0.004) & 0.328 (0.008) & 0.161 (0.025) & -0.005 (0.020) & 0.117 \\
Liver (Disorder) & 0.151 (0.001) & 0.016 (0.03) & 0.01 (0.006) & -0.012 (0.01) & 0.124 (0.006) & -0.084 (0.009) & -0.06 (0.022) & 0.056 (0.005) & 0.099 \\
Liver (Healthy) & 0.94 (0.006) & 0.033 (0.022) & 0.053 (0.004) & 0.07 (0.014) & 0.112 (0.01) & 0.062 (0.018) & 0.047 (0.01) & 0.065 (0.002) & 0.073   \\
Sonar (Mines) & 0.42 (0.011) & 0.563 (0.006) & 0.511 (0.009) & 0.396 (0.039) & 0.351 (0.008) & 0.356 (0.031) & 0.255 (0.008) & 0.133 (0.019) & 0.672\\
Sonar (Rocks) & 0.215 (0.005) & 0.147 (0.02) & 0.045 (0.015) & 0.11 (0.009) & 0.108 (0.014) & 0.151 (0.027) & 0.166 (0.032) & 0.054 (0.023) & 0.336 \\
Average  & 0.285 & 0.274 & 0.237 & 0.233 & 0.275 & 0.23 & 0.2 & 0.128 & \textbf{0.277} \\
\noalign{\smallskip}\hline
\end{tabular}
\end{adjustbox}
\end{table*}

All our experiments use five-fold cross-validation to split target class. 
To compose the complete test set we use the fold used as a test, together with all the outliers samples.
Finally, we repeat the approach twenty times taking means and variance to compare our results with other models.
In our experiments we use six numerical datasets from UCI repository (see. Table.~\ref{tab:datasets}) and a novel dataset of Covid-19 that contains chest X-ray or CT images of positive/negative patients (see Fig.~\ref{fig:covid-19_dataset}).
In order to have the right comparison criterion in Covid-19 dataset, we extract confusion matrix as reported in Tab. \ref{tab:conf_matrix_template} and compute the precision metric described in Eqs.~\ref{precision} and \ref{negative precision}. 
These last measures are needed to compare the class accuracy of the proposed OCdmst and the Resnet18 used in our experiments.

\begin{equation}\label{precision}
    {\displaystyle \mathrm {PPV} ={\frac {\mathrm {TP} }{\mathrm {TP} +\mathrm {FP} }}}
\end{equation}{}

\begin{equation}\label{negative precision}
{\displaystyle \mathrm {NPV} ={\frac {\mathrm {TN} }{\mathrm {TN} +\mathrm {FN} }}}
\end{equation}{}

\begin{table}[]
\caption{\textbf{Confusion matrix template. In x-axis predicted values, y-axis true labels value.}}
\label{tab:conf_matrix_template}
\begin{tabular}{c|c|c|c|c}                                                              
\multicolumn{2}{c}{}&\multicolumn{2}{c}{True labels}&\\
\cline{3-4}
\multicolumn{2}{c|}{}&\multicolumn{1}{c|}{Positive}&\multicolumn{1}{c|}{Negative}&\multicolumn{1}{c}{}\\
\cline{2-4}                                                                                 
\multirow{2}{*}{Predicted}& Positive & TP & FP &\ding{214} PPV\\
\cline{2-4}                                                                                 
& Negative & FN & TN & \ding{214} NPV\\
\cline{2-4}
\end{tabular}
\end{table}{}

Like for OCSVM and many others models, we need to do fine-tuning to find good parameters and obtain good results.
Considering the lack of MST due to his high computational time, it is not easy to find optimal parameters. 
Therefore, we use BFS with depth $d \in \{1, ..., 7\}$ for all UCI dataset (see details in Tab.~\ref{tab:datasets}), $d \in \{2, ..., 11\}$ only for Covid-19 dataset, a growth rate $k \in \{5,6,7,8,9,10,20\}$ and a variable $\beta \in \{1.05, 1.1, 1.5\}$ to shift the logistic function (see Fig.~\ref{fig:logistic_function}).
Furthermore, in fine-tuning parameters, we set the size of small MST from the $\frac{length\_target\ class}{4}$ to $\frac{length\_target\ class}{2}$.

\begin{table}
\centering
\caption{\textbf{Number of features, instances for all the datasets used in our experiments.}}
\label{tab:datasets}
\begin{adjustbox}{max width=0.7\columnwidth}
\begin{tabular}{l|cc} 
\hline
Datasets            & Features      & Instances   \\ \hline
Sonar                          & 60            & 208            \\ \hline
Liver                          & 6             & 345            \\ \hline
Breast cancer (Wisconsin)      & 9             & 699           \\ \hline
Diabetes                       & 8             & 768            \\ \hline
Glass                          & 9             & 214            \\ \hline
Heart                          & 13            & 297            \\ \hline
\end{tabular}
\end{adjustbox}
\end{table}

\subsection{First experiment}
In this section, we compare our OCdmst with others one-class classifiers more recent in the literature showing the results in Tab.~\ref{tab:mcc_comparison}.
For clarity, in Liver dataset we use the last field for classification. 
Also if it does not represent the presence/absence of a liver disorder, we have to use it to obtain the comparison with the other papers.
In the Heart dataset (Cleveland) we distinguish values 1,2,3,4 as the presence of disease, 0 otherwise and removed instances with missing values in according to the experiments in the literature.
Results demonstrate the rightness of our model and as reported in Tab.~\ref{tab:mcc_comparison} we outperform the state-of-the-art in terms of global mean for many classifiers (except FS-EOCC) and for many datasets.

\begin{figure}
    \centering
         \includegraphics[width=0.8\columnwidth]{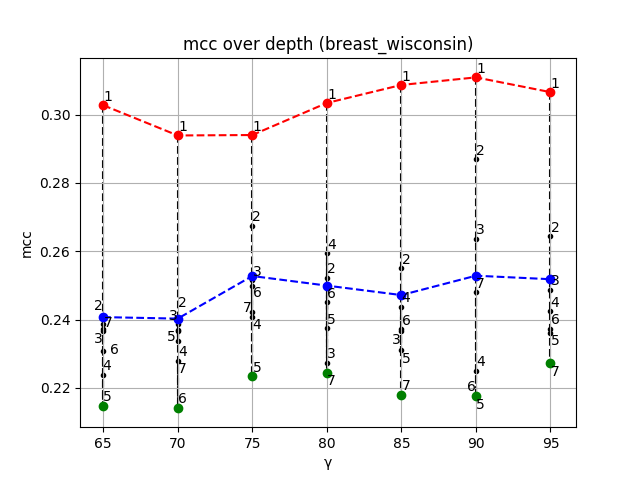}\\
         \includegraphics[width=0.8\columnwidth]{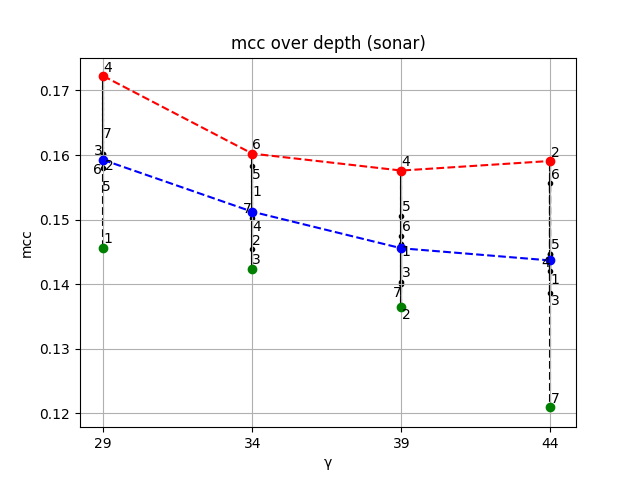}   \\
         \includegraphics[width=0.8\columnwidth]{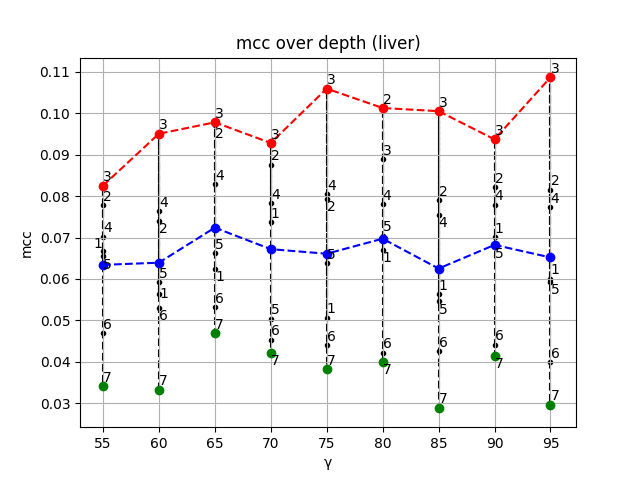} 
    \caption{\textbf{Analysis using different parameters on UCI datasets. In x-axis the size of small mst and y-axis the depth used in BFS search. Red line and blue line represent the related max and standard deviation of MCC accuracies.}} 
    \label{fig:gamma_analysis}
\end{figure}

\begin{table}
\centering
\caption{\textbf{Performance comparison using classical MST\_CD and MST\_CD\_gp with our OCdmst.
Results obtained show a clear advantages using a dynamic decision boundary. In bold our average result.}}
\label{tab:our_comparison}      
\begin{adjustbox}{max width=\columnwidth}
\begin{tabular}{l|c||cccc}
\hline\noalign{\smallskip}
Dataset & MST\_CD & $\gamma$ & $d$ & MST\_CD\_gp & \textbf{OCdmst}  \\
\noalign{\smallskip}\hline\noalign{\smallskip}
Breast (Benign)   & 0.894 (0.002) &134 &3 & 0.475 & 0.774  \\
Breast (Malignant) & -0.015 (0.011) &95 &1 & 0.183 & 0.204  \\
Diabetes (Absent)  & -0.015 (0.004) &132 &9 & 0.043  & 0.066  \\
Diabetes (Present) & 0.206 (0.003) &82 &7 & 0.152  & 0.178  \\
Glass (Float)  & 0.422 (0.005) &33  &8 & 0.401  & 0.535  \\
Glass (NoFloat) & 0.267 (0.017) &19 &3 & 0.154 & 0.238  \\
Heart (Present)  & 0.071 (0.005) &64 &1 & 0.0196 & 0.037 \\
Heart (Absent)  & 0.161 (0.025) &40 &6 & 0.091 & 0.117 \\
Liver (Disorder)  & -0.06 (0.022) &85 &2 & 0.0713 & 0.099 \\
Liver (Healthy)  & 0.047 (0.01) &66 &2 & 0.069 & 0.073   \\
Sonar (Mines) & 0.255 (0.008) &43 &9 & 0.66 & 0.672 \\
Sonar (Rocks)  & 0.166 (0.032) &24 &9 & 0.153 & 0.336 \\
Average        & 0.2 & & & 0.206 & \textbf{0.277} \\
\noalign{\smallskip}\hline
\end{tabular}
\end{adjustbox}
\end{table}

\subsection{Second experiment}
In the second experiment, we analyze our model with MCC accuracies in y-axis and the depth from 1 to 7 for each size of small MST (x-axis) (see Fig.~\ref{fig:gamma_analysis}, \ref{fig:gamma_analysis_covid}). 
Depth of BFS and size of MST are useful parameters to set up our model making as less as possible wrong classifications.

\subsection{Third experiment}
In the third experiment, we demonstrate by experimental results the effectiveness of our model with few data and compare it with a neural network (Resnet18). 
We simulate the scenario in which we have too few data to deny common neural networks to make right discrimination of different concepts (two classes in our case). 
Therefore we apply a 2-Fold cross-validation considering all data available from Covid-19 dataset and we extract the deep features for each model trained (Resnet18). 
All images have been re-scaled to 256x256 pixels and transformed to gray-scale.
Learning rate and batch-size are set to 0.001 and 10 respectively.
As evaluation metric we do not consider the total accuracy but we evaluate the accuracy for each class (Precision). 
Then, we use the deep features as input to our OCdmst using the dataset split in the previous step. 
The results reported in Tab.~\ref{tab:covid_results} show the flexibility of our model to operate also with

\begin{figure}
    \centering
    \includegraphics[width=35mm]{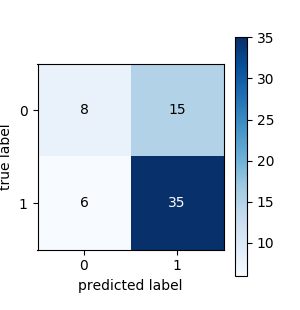}
    \includegraphics[width=35mm]{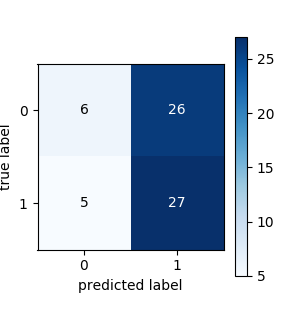}
    \caption{\textbf{Confusion matrix of OCdmst using deep features extracted by Resnet18. 
    We use k-fold cross-validation using $k=1$ (image on the left) and $k=2$ (image on the right).
    Numerical results are showed in Tab. \ref{tab:covid_results}.}}
    \label{fig:confusion_matrix}
\end{figure}

Resnet18 learned features using cross-entropy loss.
As showed in Tab.~\ref{tab:covid_results}, Resnet18 does not offer right prediction to recognize instances from negative class. 
This results is reasonable considering we have too few data with negative labels. 
Even when we used features extracted by Resnet18 to our model, but OCdmst is able to recognize value from both classes and overcome the right prediction also on positive class.
The confusion matrix showed in Fig. \ref{fig:confusion_matrix} and accuracies for each class demonstrate the capability of OCdmst to operate also in a scenario with few data, as well as overcome deep neural networks. This last result is plausible because it represents one of the weaknesses of deep models.

\section{Conclusion}
In machine learning, non-uniformly sampled data is a problem that normally leads the predictive models to make misclassifications.
To contrast this issue, in this paper, we propose a dynamic boundary approach for one-class classification able to handle their own boundary also in a non-uniformly sampled data. 
The effectiveness and solidity of our approach have been compared with many one-class classifiers and for many of them, we have passed the state of the art.
We believe that our model can be improved in terms of computational time, in order to be applied also in computer vision.

\bibliographystyle{main}
\bibliography{main}


\end{document}